# Semi-Supervised Hierarchical Semantic Object Parsing

Jalal Mirakhorli
Department of Electrical Engineering
Amirkabir University of Technology
jalalmiry@gmail.com

Hamidreza Amindavar
Department of Electrical Engineering
Amirkabir University of Technology
hamidami@aut.ac.ir

*Abstract*—Models based on Convolutional Neural Networks (CNNs) have been proven very successful for semantic segmentation and object parsing that yield hierarchies of features. Our key insight is to build convolutional networks that take input of arbitrary size and produce object parsing output with efficient inference and learning. In this work, we focus on the task of instance segmentation and parsing which recognizes and localizes objects down to a pixel level base on deep CNN. Therefore, unlike some related work, a pixel cannot belong to multiple instances and parsing. Our model is based on a deep neural network trained for object masking that supervised with input image and follow incorporates a Conditional Random Field (CRF) with end-to-end trainable piecewise order potentials based on object parsing outputs. In each CRF unit we designed terms to capture the short range and long range dependencies from various neighbors. The accurate instance-level segmentation that our network produce is reflected by the considerable improvements obtained over previous work at high $AP_r$ thresholds. We demonstrate the effectiveness of our model with extensive experiments on challenging dataset subset of PASCAL VOC2012.

## I. INTRODUCTION

Scene & object parsing is a fundamental task in computer vision, Its goal is to assign one of many pre-defined category labels to each pixel in an image. That use wide applications in field of autonomous vehicles, robot navigation, medical imaging systems (comprised of x-ray CT, MRI, ultrasound, nuclear medicine) and virtual reality. However, it remains a challenging problem since it requires solving segmentation, classification and detection simultaneously. Top performing image classification architectures usually involve very deep CNN trained in a supervised fashion on a large datasets [1-2] and have been shown to produce generic hierarchical visual representations that perform well on a wide variety of vision tasks. However, these deeper CNNs heavily reduce the input resolution through successive applications of pooling or subsampling layers. To address this issue, we used Convolutional Neural Network (CNN) to learning image representations and applied for class base object masking as in "Mask R-CNN" [3]. However, CNNs can capture high-level context information learned at top layers that have large Receptive Fields (RFs) that supervise with RF superpixel in the feature map, because first layers in CNNs are not exposed to valuable context information when they learn features and could be last layers augment via superpixels RFs, as instance segmentation. To effectively address the objects parsing problem, then employ Conditional Random Fields (CRFs) as a post-processing step to object parsing in instances, this method allows models with highly expressive feature, without worrying independencies, in CRF don't need to model distribution over variables and can train a discriminative classifier to improve performance.

In this paper, we aim at the efficient application of CNN&CRF to building a practically fast instance object parsing system with decent prediction accuracy, retrieve contextual information from images. We propose to extend the Mask Regional proposal CNN architecture [3], originally designed for image classification and masking, to deal with the more ambitious task of semantic segmentation.

Many recent object parsing methods, are based on the Detection and Segmentation approach [4]. These methods [4-5] generate object proposals [6-7], classify each proposal into an object category, and then employ a bounding box processing into a segmentation of the primary object the proposal contains. However, because this three step proposal generating in whole image, object detection in all regional proposals and bounding boxing in the first algorithm take more time to process an image. We propose a different approach to object detection and parsing, where we integrate all processing for object masking and instance semantic segmentation together before object parsing in the scene via [3] method while improve with superpixel pooling layer, And then a Conditional Random Field (CRF) can use cues from the output of an object masking, this CRF can be joint as a layer of a neural network. Our contributions can be summarized in the following three aspects. 1) A novel Convolutional Neural Network model is proposed for object detection and masking, which jointly optimizes over superpixels for, object masking and relation prediction. 2) CRF modeling for improvement object parsing. 3) Extensive experiments on public benchmark demonstrate the superiority of our model over other state-of-the-art object parsing approaches.

## II. RELATED WORK

Object instance and parsing segmentation have most challenging problems in computer vision, which aims to decompose objects into their semantic parts, has been addressed by numerous works that most of which have concentrated on parsing humans. However, most researches works have parsed objects at an instance level, but work rather category level.

Recently, Convolutional and Recurrent neural networks based methods achieved great success in these tasks. Several recent studies [8-9] made one of the earliest attempts at applying Graph Long Short-Term Memory (LSTM) in features produced by CNNs to object parsing. [10] Focuses on semantic understanding of human bodies for human parsing, which has several appealing properties by CNN and jointly segmentation and default structure losses. Liang et al. [11] proposed to use two separate convolutional networks to predict the template coefficients for each label mask and their corresponding locations, respectively, focused on datasets. Most of the aforementioned works using Fully Convolutional Network (FCN) [12] that trained with part labeling as in [13] and aren't based on object classes. However, they need post-processing to extract object bounding box and instance segmentation as demonstrated in [8]. In these works form priors of the objects have been encoded using pictorial structures [14], Conditional Random Fields (CRFs) [15-17] and LSTMs [8-9]. [18] Developed proposal-free method that exploit the capability of global context information by different-region-based context aggregation through via pyramid pooling module together with the proposed Pyramid Scene Parsing Network (PSPNet). this global prior representation is effective to produce good quality results on the scene parsing task, while PSPNet provides a superior framework for pixel-level prediction, that extend the pixel-level feature to the specially designed global pyramid pooling one. The local and global clues together make the final prediction more reliable. In the following, "ICNet" [19] method proposed a compressed-PSPNet-based image cascade network that incorporates multi-resolution branches under proper label guidance to reducing a large portion of computation for pixel-wise label inference. In [20] used unary terms that produced by employing a 2D articulated model by [21] which predicts the main keypoints or exploits the statistical dependency between the location on the desired object, also use eSIFT and eMSIFT proposed by [22] for region description.

Most prior works are very complicated due to several pre-processing and post processing steps. Various methods of instance segmentation have also involved modifying object detection system to output segments instead of bounding boxes [23-5]. However, they consider each one of detection independently. In addition, combining independent steps is not an optimal solution. In contrast, we make a directly predicts pixelwise instance location maps and uses a simple masking [3, 24] technique to generate instance-level segmentation results. Our network predicts the instance number in a totally data-driven way by the trained network [3, 23], which can be naturally scalable and easily extended to other instance-level recognition tasks. More details will be provided in following sections.

## III. MODEL DESCRIPTION

Our network (Fig. 1) consists of two sub-networks: an initial object masking segmentation, sub-network 1, that improvement with supervises superpixel pooling layer, and an object parsing, sub-network 2. As both of these sub-networks are differentiable, they can be integrated into a single network and trained jointly. However, two training steps are employed to train the networks. First, we train the "Mask R-CNN" [3] without superpixel pooling layer to produce object masking. Then, the whole sub-network 1 is fine-tuned based on the pertained model with superpixel layer to produce final instance segmentation results.

We used output of first sub-network with the superpixel layer produced, within the second Mask R-CNN network and Conditional Random Field (CRF), respectively, to compute object parsing over instances. We perform mean field inference in this random field to obtain the Maximum a Posteriori (MAP) estimate, which is our labeling.

### A. Object Masking

Our proposed model builds on top of "Mask R-CNN" [3] and improves it to address the task of object labeling. The Mask R-CNN efficiently detects objects in an image while simultaneously generating a high-quality segmentation mask for each instance. This method extends Faster R-CNN [25] by adding a branch for predicting an object mask in parallel with the existing branch for bounding box recognition. The mask branch is a small FCN applied to each Region of Interest (RoI), predicting a segmentation mask in a pixel-to-pixel manner. In principle Mask R-CNN is an intuitive extension of Faster R-CNN, yet constructing the mask branch properly is critical for good results. In this model [3] define a multi-task loss on each sampled RoI as $L_{(total)} = L_{(classification)} + L_{(box)} + L_{(mask)}$. The classification loss and bounding-box loss are identical. The mask branch encodes K binary masks of resolution m × m, one for each of the K classes and applies a per-pixel sigmoid, and we define $L_{(mask)}$ as the average binary cross-entropy loss , then the $L_{(mask)}$ is only defined on the k-th mask without competition among classes [3], in Mask R-CNN model used RoIAlign layer instead of RoIPool [26] to remove the harsh quantization.

According to the above-mentioned, we are going to improve this method by using Superpixel Pooling Network (SPN), defined in [27], and CRF. We can employ superpixel segmentation as a pooling layout to reflect low-level image structure for learning and inferring objects instance segmentation in a semi-supervised setting. SPN takes two inputs for inference: an image and its superpixel map. Given an input image, our network extracts high-resolution feature maps using encoder and then apply several layers of upsampling layers, the superpixel pooling layer aggregates features inside of each superpixel by exploiting an input superpixel map as pooling layout. Then relevance of superpixels to instances categories is obtained by training with discriminative loss, as shown in Fig. 1, we put additional branch of global average pooling for regularization, this is branch of before upsampling processing, to prevent undesirable training noises introduced by superpixels [27]. To implementation superpixel pooling layer, we require to two inputs, a superpixel map with size M x N that directly create from input image and a feature map size K x M x N that produced in last layer of processes object detection. With pooling the features (here average-pooling) belonging to the same superpixel, where K is the feature map depth or channels. Then pooled feature vector for i-th superpixel in average-pooling layer given by:

$$\bar{Z}_i = \frac{1}{k_i} \sum_j \sum_k I(P_i^k \epsilon\, r^j)\hat{z}^j \qquad (1)$$

Where $p_i$ is i-th superpixel and k, number of pixel in $p_i$. $r^j$ And $\hat{z}^j$ represent the receptive field and the feature vector of the j-th location in $\hat{z}$, respectively. $I(P_i^k \epsilon\, r^j)$ Is an indicator function that is 1 if the center of $r^j$ is closer to $P_i^k$ than those of any other receptive fields are, and 0 otherwise. Then obtained global average pooling overall superpixels, computed as $\tilde{z} = 1/N \sum_i \bar{z}_i$, where N indicates the number of superpixels. The image-level feature vector $\tilde{z}$ is then classified by the fully connected layer following the superpixel layer, and the result is fed to the loss function £. The above algorithm is used for instance segmentation and object parsing, respectively, in the both sub-networks.

B. CRF

We boost our model's ability to capture fine details by employing a Conditional Random Field (CRF). At the input to our second sub-network, we assume that we have objects masking and superpixels that produce in the last layer of the first sub-network. The problem of object parsing can then be thought of as assigning every pixel to either a particular instance object parsing. For this work, we define a multinomial random variable, V, at each of the N pixel in the instance objects, V= [$V_1$ $V_2$ ....$V_n$]. Each variable at pixel i, $V_i$, is assigned a label corresponding to every parse of objects. This label set, {0, 1, 2, ..., n} changes for each individual object, the number of parse labeling that varies and corresponding with number of possible pieces of desired object.

We formulate a Conditional Random Field over our parse variables, V, which consists of unary and pairwise energies, unary potentials represent the confidence of parse hypotheses based on the local object evidence, while pairwise potentials model spatial arrangement of parses in the object. The energy of the assignment v to all the variables, V, is:

$$E(V = v) = \sum_i \psi_i(v_i) + \sum_{i<j} \psi_{ij}(v_i, v_j) \qquad (2)$$

The unary potential $\psi_i(v_i)$ specifies the energy cost of assigning label, $v_i$, to pixel i. the second Mask R-CNN outputs a probability estimate of each pixel containing each parse label. Denoting the output of the Mask R-CNN for pixel

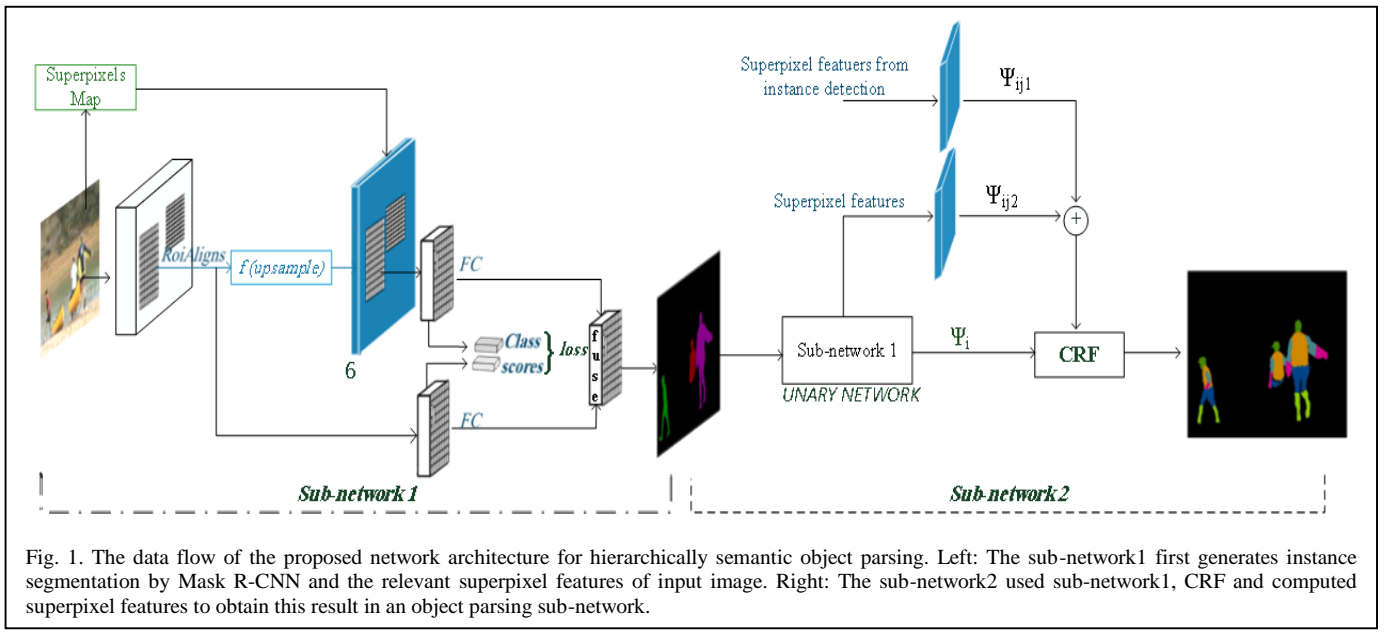

Fig. 1. The data flow of the proposed network architecture for hierarchically semantic object parsing. Left: The sub-network1 first generates instance segmentation by Mask R-CNN and the relevant superpixel features of input image. Right: The sub-network2 used sub-network1, CRF and computed superpixel features to obtain this result in an object parsing sub-network.

i and label $v_i$ as $z_{i:v_i}$, the unary potential is:

$$\psi_i(v_i) = -\alpha_u \log(z_{i:v_i} + \varepsilon) \quad (3)$$

Where $\alpha_u$ is a parameter controlling the impact of the unary potentials, and $\varepsilon$ is introduced to avoid numerical problems [28].

The pairwise potential $\psi_{ij}(v_i, v_j)$ specifies the energy cost of assigning label $v_i$ to pixel i while pixel j is assigned label $v_j$. Introducing pairwise terms in our model enables us to take dependencies between output data into account. Our pairwise network generates the similarity matrices that are used in the pairwise term of the energy function (1). In this work, we used the feature vector of superpixels that produced in the last layer of the both sub-networks for the pairwise term. Therefore, neighboring superpixels are typically connected via a (contrast-sensitive) Potts model encouraging smoothness of the labels. For objects parsing, we want these connections to act on a longer range. Our goal is to encourage superpixels that are similar in appearance to agree on the label, even though they may not be neighbors in the image, in fact, the pairwise term consists of densely-connected Gaussian potentials [29] and encourages appearance and spatial consistency. These labels were later assigned to pixels belonging to its superpixel. With the total of all similarity features between each pair of superpixels, we can measure how well two superpixels fit each other. Then train a logistic regression to predict if the two superpixels should have the same label or not. For this work, we minimized the spanning tree algorithm using the similarity matrix and use the top 8 edges to connect 8 pairs of superpixels feature in each instance object to alleviate slow down inference processing in this work, we compute a pairwise potential between each connected pair, superpixels, $\Phi_{i,j}^{similarity}$ [20], to define similarity classifier. More details are in section D.

### C. Inference of instance CRF

The energy of the full model define the sum of two types of energies encoding unary and pairwise potentials that depend on the pixel and superpixel labeling, respectively, also two components of energy function are differentiable with respect to their input and parameters. Follow describe how this energy can maximized during inference, since the energy in (2) characterizes a Gibbs distribution, the Maximum A Posteriori (MAP) estimate of our CRF is computed as the final labeling produced by our network. We perform the iterative mean-field inference algorithm to approximately compute the MAP solution by minimizing (2) via a backpropagation algorithm. As shown by Zheng et al. [30], this can be formulated as a Recurrent Neural Network (RNN), allowing it to be trained end-to-end as part of a larger network. However, number of object parsing is variable in object then the label space of the CRF is dynamic. Therefore, unlike [30], the parameters of our CRF are not class-specific to allow for this variable number of channels.

### D. Loss function and network training

In this work, we have two sub-networks that have been trained using different processing. For train the first sub-network, Mask R-CNN with superpixel pooling layer. As shown by [27], the SPN is learned with classification losses and since multiple objects of different classes may appear in an image, therefore, loss function defined as the sum of C binary classification losses. This model used scale variations of objects to increase the accuracy of the detection and follow the approach of [31] that randomly resizes images in the input mini-batch during training. Note, this multi-scale approach only fit to superpixel layer. As shown in (1), SPN assigns a feature vector to each superpixel through the superpixel layer. After inference processing SPN, when the class scores of the individual superpixels are computed, we obtain a tensor in each channel corresponds to an activation map for the associated class. Thus used multi scale input image in training phase then aggregated activation tensors by max-pooling and produced Superpixel-Pooled Class Activation Map (SP-CAM) [27]. Finally, the feature vectors of each superpixels candidate and hold for compute similarity matrix in second sub-network.

The second sub-network included Mask R-CNN, with superpixel layer, which trained for object parsing while improves by CRF, we employs this structure to each of the objects extracted. As already mentioned in (2), the energy function has two components, unary&pairwise potentials. We use (3) as unary potential, $\psi_i(v_i)$, where $(z_{i:v_i} + \varepsilon)$ is the label assignment probability at pixel i as computed by Mask R-CNN which improve via a superpixel layer.

The pairwise potential is $\psi_{ij}(v_i, v_j) = \sum_{n=1}^{k} \omega_n * k^n(f_i, f_j)$, there is one individual pairwise potential for each pair of the superpixel in the desired object, i.e. the model's factor graph is fully connected. Each $k^n$ is the Gaussian kernel depends on features, Z, extracted for superpixles i, j and also is weighted by parameter $\omega_n$. Therefore, we adopt bilateral position and feature extract form the first sub-network terms, In particular, as in [29], we use the following expression for pairwise term:

$$\psi_{ij}(v_i, v_j) = \mu(v_i, v_j) \sum_{n=1}^{k} \omega_n * k^n(f_i, f_j) =$$
$$\mu(v_i, v_j) \left[ w_1 \exp\left(-\frac{\|p_i - p_j\|^2}{2\sigma_\alpha^2} - \frac{\|z_i - z_j\|^2}{2\sigma_\beta^2}\right) + w_2 \exp\left(-\frac{\|p_i - p_j\|^2}{2\sigma_\gamma^2}\right) \right] \quad (4)$$

Where $\mu(v_i, v_j) = 1$ if $v_i \neq v_j$, and zero otherwise and the first kernel depends on both superpixel positions and superpixel features, and the second kernel only depends on superpixel positions. The hyper parameters $\sigma_\alpha$, $\sigma_\beta$ and $\sigma_\gamma$ control the "scale" of the Gaussian kernels. This model is amenable to efficient approximate probabilistic inference [29]. The message passing updates under a fully decomposable mean field approximation $b(v) = \prod_i b_i(v_i)$ can be expressed as convolutions with a Gaussian kernel in feature space [32]. As shown in Fig. 1, we used superpixel layers created in both sub-network for pairwise term in CRF, then the energy of the full model is the sum of the three term of energy unary and pairwise potentials the depend on superpixel labeling, the full model is:

$$E(Y \mid I) = E_{unary}(Y) + E_{similarity1}(Y \mid I) + E_{similarity2}(Y \mid I) \quad (5)$$

## IV. EXPERIMENT

This section describes the experimental analysis and results, followed by visual inspection of the quality of the obtained label maps on PASCAL-Person-Part validation set.

In the first step, The Mask-RCNN parts of our model were initialized with the trained model provided by the authors of [3], without the superpixel layer and CRF, for object detection and object parsing, respectively. We then integrate proposed layers and train both sub-network separately. We evaluate our first sub-network using the challenging PASCALVOC 2012 segmentation dataset [33], which consists of 20 object classes and one background class. We also augment the training set with the additional annotations provided by Berkeley segmentation data for PASCAL VOC 12.

For evaluated second sub-network, object parsing, we used the human parsing performance of the second Mask R-CNN on The public PASCAL-Person-part dataset that is a fine-grained part segmentation benchmark collected by Chen et. al. [34] from PASCAL VOC 2010 dataset. It contains the detailed part annotation for each person. Following [8, 13], the annotations are merged into six person parts classes, (i.e. Head, Torso, Upper/Lower Arms and Upper/Lower Legs) and one null class. Totally, 1,716 images are used for model training and 1,817 for test.

We used the standard $AP_r$ metric [4] for evaluating instance-level segmentation, and mean intersection-over-union criterion and pixel-wise accuracy to evaluate object parsing network and other comparative methods. The $AP_r$ metric computes the mean average precision under different Intersection over Union (IoU) scores between the predicted and ground truth segmentation. The number of superpixels, image scaling and training parameters used in both the sub-network are similar to [27].the SLIC [35] over input image and object instance segmentation method used to generate superpixels. Fig. 2, provides a visual comparison of the proposed approach with Ground truth.

We compare our method with four baselines .As shown in Table 1, the proposed structure provides considerably better prediction for the semantic parts such as torso, u-legs and background than baseline methods. The evaluated metric, AP, demonstrated which additional superpixel layer inside masking network can improve the instance segmentation around 4% than original method.

## V. CONCLUSION

In this work, we proposed a novel semi-supervised network to address semantic object parsing task. This method to extract instance-wise object part parsing is based on superpixel pooling layer. The proposed network consists of Mask R-CNN network, CRF and a superpixel map to improve both outputs of CRF inference and the masking network. We showed proper and hierarchical combination for feature extraction can improve next level networks, so, continuing this approach makes smaller semantic parts, just like extracting data in a pyramid form. The use of a supepixel pooling layer in this work demonstrated substantially improved quality at object's edge, reduce side effects introduced noisy, for classes with small number of data, and do not require explicit object part annotations.

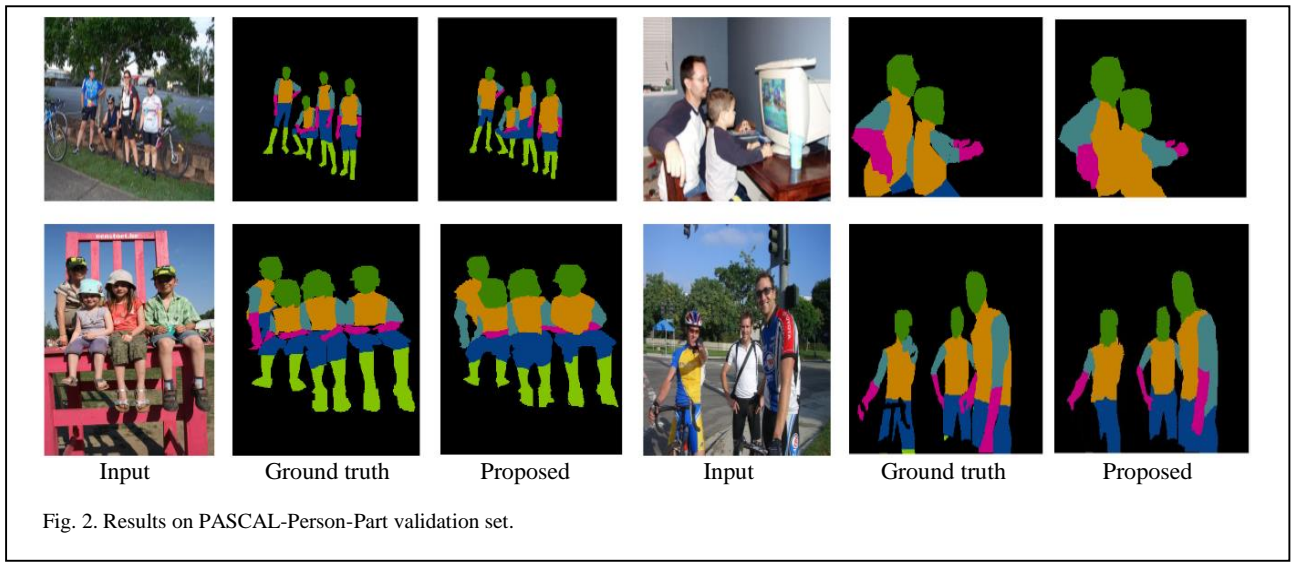

Input     Ground truth     Proposed     Input     Ground truth     Proposed

Fig. 2. Results on PASCAL-Person-Part validation set.

TABLE 1
Comparison of person part segmentation performance with five state-of-the-art methods
On the PASCAL Person-Part dataset.

| Method | head | torso | u-arms | l-arms | u-legs | l-legs | Bkg | Avg |
|---|---|---|---|---|---|---|---|---|
| DeepLab-LargeFOV [36] | 78.09 | 54.02 | 37.29 | 36.85 | 33.73 | 29.61 | 92.85 | 51.78 |
| DeepLab-LargeFOV-CRF | 80.13 | 55.56 | 36.43 | 38.72 | 35.5 | 30.82 | 93.52 | 52.95 |
| HAZN [37] | 80.79 | 59.11 | 43.05 | 42.76 | 38.99 | 34.46 | 93.59 | 56.11 |
| Attention [13] | 81.47 | 59.06 | 44.15 | 42.5 | 38.28 | 35.62 | 93.65 | 56.39 |
| Graph LSTM [8] | **82.69** | 62.68 | **46.88** | **47.71** | 45.66 | **40.93** | 94.59 | 60.16 |
| proposed | 81.06 | **63.92** | 43.65 | 46.87 | **46.93** | 38.85 | **95.07** | 59.47 |